\documentclass[a4paper,conference]{IEEEtran}

\usepackage{microtype}
\usepackage{booktabs} 
\usepackage{amsfonts}
\usepackage{bbm}
\usepackage{dsfont}
\usepackage{adjustbox}
\usepackage{multirow}
\usepackage{amsmath}

\usepackage{subfig}

\DeclareMathOperator*{\argmax}{argmax} 

\usepackage{hyperref}

\begin{document}

\title{Performance Prediction Under Dataset Shift}






\author{\IEEEauthorblockN{Simona Maggio}
\IEEEauthorblockA{Dataiku\\
simona.maggio@dataiku.com}
\and
\IEEEauthorblockN{Victor Bouvier}
\IEEEauthorblockA{Dataiku\\
victor.bouvier@dataiku.com}
\and
\IEEEauthorblockN{Léo Dreyfus-Schmidt}
\IEEEauthorblockA{Dataiku\\
leo.dreyfus-schmidt@dataiku.com}}



\maketitle

\begin{abstract}

ML models deployed in production often have to face unknown domain changes, fundamentally different from their training settings. Performance prediction models carry out the crucial task of measuring the impact of these changes on model performance.
We study the generalization capabilities of various performance prediction models to new domains by learning on generated synthetic perturbations. Empirical validation on a benchmark of ten tabular datasets shows that models based upon state-of-the-art shift detection metrics are not expressive enough to generalize to unseen domains, while Error Predictors bring a consistent improvement in performance prediction under shift. We additionally propose a natural and effortless uncertainty estimation of the predicted accuracy that ensures reliable use of performance predictors. Our implementation is available at \textnormal{\url{https://github.com/dataiku-research/performance\_prediction\_under\_shift}}.

\end{abstract}




\section{Introduction}
\label{intro}

To ensure safe and robust deployment of ML models in production, it is critical to assess the reliability of its predictions on unknown data. Data shift, \textit{i.e.} when incoming data is structurally different from the training data, is a typical situation where model performance may drop significantly. One can detect such scenario leveraging shift detector models. However, this does not quantify the extent of the expected drop in model performance.

As all shifts are not equally harmful for a trained model, we address the understudied problem of predicting model accuracy drop on shifted data. A traditional approach builds dataset-level features extracted from shift detectors, which a secondary model, named the \textit{performance predictor} (sometimes referred to as meta-model \cite{Elder20}), then uses to predict the model performance drop. This approach raises several challenges. First, fitting a performance predictor requires labeled domains that are representative enough of real-world shifts \cite{Elsahar2019, Redyuk2019}. As collecting labeled domains is difficult, it is more feasible to train the performance predictor on synthetically shifted datasets, applying label-preserving perturbations \cite{Redyuk2019}. Second, computing dataset-level features from the shift detector is highly expensive, limiting its practicability for real-time prediction. A recent approach, \textit{Average Threshold Confidence} (ATC) \cite{Garg2022}, predicts model accuracy as the fraction of incoming data for which model confidence exceeds a tuned threshold. In spite of being computationally light, ATC assumes that wrong predictions are tied to specific model confidence levels, but there is no guarantee that this assumption holds under distribution shift, where the calibration of the model confidences does not translate \cite{Ovadia2019}.

Existing works present heterogeneous experimental settings where shifts are either unrealistic or present small diversity. This limitation, together with the lack of a standard benchmark for evaluating performance predictors, makes it difficult to determine the best performing and robust approach. To this purpose, this work provides an in-depth comparison between competing and complementary approaches to predict performance drop. In particular, we study three methods; learning from tailored features from drift detection tasks \cite{Elsahar2019, Redyuk2019}, a model-free approach based on the confidence values of the model \cite{Garg2022}, and the aggregation of sample-wise error prediction \cite{Elder20}. As we are care for the robustness of performance predictors under dataset shift, we focus on the ability of these models to generalize to unseen domain changes, either \textit{synthetic} or \textit{natural}, where the latter is obtained by extracting subpopulations from the original datasets without any perturbation.

We summarize our contributions as follows:  

\begin{enumerate}
\item We show that Error Predictors trained on synthetic shifts yield consistent improvement to the performance prediction task under distribution shifts. Our results especially highlight that approaches based on drift detection metrics generalize poorly under shift, suggesting that such metrics are appropriate for detection but not to estimate how shift impacts performance.

\item We propose a benchmark over ten tabular datasets against state-of-the-art methods together with appropriate metrics accounting for sampling uncertainty. This provides empirical evidence of the superiority of Error Predictors.  

\item We present a natural and effortless uncertainty estimation of accuracy from performance predictors that proves crucial for trustworthy use of performance predictors.
\end{enumerate}
We provide in Section \ref{sec:related_works} an overview of the prior works and Section \ref{performance_predictors} details the methods we compare. Section \ref{sec:validating_pp} describes our protocol, while Section \ref{sec:experiments} summarizes our empirical investigation. Section \ref{sec:conclusion} concludes the paper.

\section{Related work}
\label{sec:related_works}
Since ML models generalize poorly when data is subject to shift \cite{DAmour2020}, estimating the performance drop is crucial for their safe deployment and thus has received a growing interest in the latest years. One can distinguish between two families of methods, whether it averages the predicted sample-wise correctness of the model, or if it predicts the performance drop from the whole unlabeled target data.

The former family relies on proxies of the sample-wise model error; by learning an error predictor \cite{Elder20}, a confidence score \cite{Chen2019}, or by using a scoring function that below a tunable threshold indicates a model error \cite{Garg2022}. The latter family relies mostly on dataset similarity \cite{Guerra2008, Schat2020}; by extracting features obtained derived from a shift detector  \cite{Elsahar2019,Redyuk2019}, or obtained from past data \cite{Talagala2019}. We compare these existing approaches in this study and propose strategies that improve their generalization to unseen domain changes. A new line of study aims to achieve unsupervised calibration on unlabeled target data \cite{Deng2021-2, Deng2021, Guillory2021}.

Fitting a predictor from dataset-level features typically requires access to a set of labeled domains with various shifts. As such domain diversity is hard to obtain in practice, prior works have investigated the roles of synthetic shifts  \cite{Redyuk2019, Elder20} for achieving this task. The present work focuses more particularly on understanding how synthetic data augmentation can improve predictions on natural shifts, as described in \cite{Taori2020} for computer vision systems.

\section{Performance Predictors}
\label{performance_predictors}

The evaluation of ML model performance on samples is only representative of their true performance under independently and identically distributed (i.i.d) settings.  If there exist various upper bounds of a model's risk under distribution shifts, starting from Ben-David's seminal work \cite{Ben-david2010}, we can strive to directly estimate it.  
Once a change of domain is detected, we can leverage various statistics and shift indicators to assess how harmful it is to the model performance. However the exact nature of the new domain being often unknown, we resort to simulating new domains by generating synthetic shifts (Subsection \ref{subsec:shifts}) and train performance drop predictor on this augmented set of datasets.  
We present and test three strategies for this task: \textit{Expert Models} leveraging model statistics and state-of-the-art shift indicators between datasets, a simple heuristic derived from the model confidence and an aggregated \textit{Error Predictor} that predicts error at the sample level.

We denote by $f: X \to Y$ a classifier with accuracy $\widetilde{acc}_{\mathcal{D}}$ on a test set $\mathcal{D}$. The set of augmented test sets with the corresponding accuracy of $f$ is denoted by $\{ \mathcal{D}_{k}, \widetilde{acc}_k \}_{k=1, \dots, K}$. We refer to these test set as \textit{target} datasets while we refer to the dataset used for training the model $f$ as the \textit{source} dataset. 

It should be noted that the impact of shift on other metrics such as precision, recall and AUC should also be examined, but we leave this to future research, while focusing here solely on the properties of the accuracy metric. 

\subsection{Expert Models}

We present two versions of \textit{Expert Models}. The first proposed in \cite{Elsahar2019} is a Logistic Regression trained on data shift detection metrics computed from the comparison of each training dataset $\mathcal{D}_k$ with the source dataset $\mathcal{D}$ such as PAD (Proxy A-Divergence), RCA (Reverse Classification Accuracy), and confidence drop. The second expert model proposed in \cite{Redyuk2019} is a Random Forest Regressor trained on percentiles of the predicted class probabilities of the primary model $f$ on the shifted sets $\mathcal{D}_k$.

For a fair comparison, we consider the same underlying model, a Random Forest Regressor, for both expert models above.


\subsection{Average Thresholded Confidence}

The Average Thresholded Confidence (ATC) \cite{Garg2022} is a model-free technique, predicting as errors all samples for which the maximum confidence or negative entropy of the primary model is below a given threshold.  
The threshold $t$ is identified as the one for which the ratio of samples with negative entropy below this threshold best approximates the true error rate on a source test set $\mathcal{D}^S$. If $f_j(x)$ denotes the confidence of the model that observation $x$ belongs to the $j-th$ class of the class set $\mathcal{Y}$, this means:
\begin{multline}
\mathds{E}_{x \sim \mathcal{D}^S}\left[\mathds{I}\left(\sum_{j\in\mathcal{Y}} f_j(x) \log(f_j(x)) < t\right)\right] = \\
\mathds{E}_{(x,y) \sim \mathcal{D}^S}\left[\mathds{I}\left(\argmax_{j\in\mathcal{Y}} f_j(x) \neq y\right)\right].
\end{multline}

\subsection{Sample-wise Error Predictor}


Different from the above performance predictor acting at the dataset level, the sample-wise \textit{Error Predictor} aims at predicting when the primary model's prediction on a sample is correct. The predicted accuracy over a dataset is then given as the mean of \textit{Error Predictor} predictions.  

This performance predictor uses the original raw features of the dataset, the base model highest class probability and the base model margin uncertainty (difference between the two highest probabilities). If a typical choice for this binary classifier is Gradient Boosting \cite{Elder20}, we show the importance of model selection in the experiment section.  

Note that in \cite{Elder20} this type of performance predictor is referred to as \textit{meta-model}, but for the sake of clarity in this work we avoid using this term, because we consider that performance prediction itself calls for a meta-learning approach, so that \textit{any} type of performance predictor is a meta-model.


\section{Validating Performance Predictors}
\label{sec:validating_pp}

Model's measure performance are subject to sampling uncertainty. For accuracy, if $\text{acc}$ denotes the true model accuracy in the limit of infinite sample, by the central limit theorem we know that the measured accuracy $\widetilde{acc}$ over an i.i.d dataset of size $n$ follows a normal distribution $\widetilde{acc} \sim \mathcal{N}(\text{acc}, \sigma)$ with $\sigma=\sqrt{\frac{\widetilde{acc} \cdot(1-\widetilde{acc})}{n}}$ \cite{Pishro2014}.

The above suggests that we should account for variation of accuracy when evaluating performance predictors.
Indeed, even when performance predictors might produce a wrong prediction, this can still be included in the confidence interval of the true accuracy at the practitioner's desired confidence level $\alpha$, in which case the performance predictor should not be penalized. This confidence interval is defined as $\widetilde{acc} \pm z_{\alpha/2} \sigma$, where $z_{\alpha/2}$ is the number of standard deviations of a Gaussian distribution spanning the confidence level $\alpha$ \cite{Pishro2014}.

In order to take into account this acceptable prediction error, in addition to standard metrics such as the absolute error,
we propose to evaluate a variant of the mean absolute error, which count as zero error when the prediction falls in the confidence interval at significance level $\alpha$, while considering the error equal to the distance to the confidence interval closest bound, when the prediction falls outside the interval. In the experiments we consider the significance level $\alpha=0.05$.

The described metric is given below, where $K_t$ is the number of datasets in the test set, $\widetilde{acc}_k$ (resp. $\widehat{acc}_k$) the measured ground-truth (resp. predicted) accuracy on the $k$-th dataset:
\begin{itemize}
\item MAE within Confidence Interval at fixed $\alpha$, $MAE_{CI_\alpha}$:
\begin{multline}
 \frac{1}{K_t}\sum_k \max\left(0, |\widetilde{acc}_k - \widehat{acc}_k| -z_{\alpha/2}\cdot\sigma(\widetilde{acc}_k)\right).
\end{multline}
\end{itemize}

\section{Direct Uncertainty Estimation for Performance Prediction}

Prediction on out-of-distribution samples is a challenging task that warrants the needs for additional uncertainty estimation. Producing this uncertainty estimate for performance predictors requires to either use specific models that directly return uncertainty estimation or to train yet another separate model, based on features extracted from the performance predictor \cite{Elder20}. In this section we present a simple yet sound approach to estimate confidence intervals for point-wise prediction of any performance predictor.

In Section \ref{sec:validating_pp} we highlighted the inherent uncertainty of the measured accuracy to motivate evaluation metrics which do not penalize prediction within the confidence interval of the ground-truth measured accuracy. A discrete version of the $MAE_{CI_\alpha}$ metric is $ACC_{CI_\alpha}$, measuring the ratio of predictions falling in the confidence intervals of the true values. This metric measures the coverage of the true accuracy intervals on the predicted values.
 
We model the predicted accuracy on a dataset by a performance predictor as $\widehat{acc} \sim \mathcal{N}(a, \hat{\sigma})$ with $\hat{\sigma}=\sqrt{\frac{\widehat{acc} \cdot(1-\widehat{acc})}{n}}$, assuming the performance predictor provides good estimates of true accuracy.
We derive from this a simple approach to estimate confidence intervals for point-wise accuracy predictions by considering the interval $\widehat{acc} \pm z_{\alpha/2}\times \hat{\sigma}$ for a given significance level $\alpha$. 
 
\begin{figure}[t]
\begin{center}
\centerline{\includegraphics[width=\columnwidth]{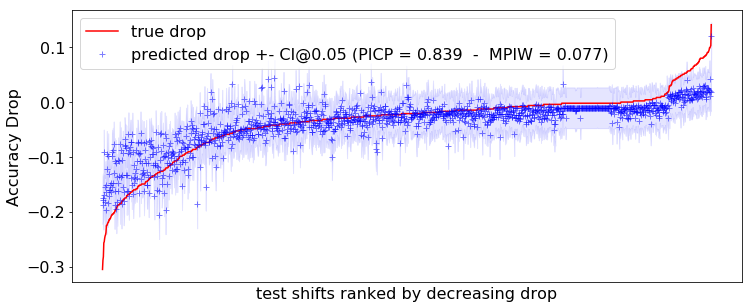}}
\caption{Confidence intervals at $\alpha=0.05$ for the \textit{Error Predictor} outcome on drift scenarios of the Video Games dataset.}
\label{fig:uncertainty_video_games}
\end{center}
\end{figure}

Figure \ref{fig:uncertainty_video_games} shows the confidence intervals found by this approach on the predictions of the \textit{Error Predictor} for several target datasets at various accuracy drops in the Video Games dataset. We report the two standard metrics for the goodness of confidence intervals: the Prediction Interval Coverage Probability (PICP) at $0.839$ and the Mean Prediction Interval Width (MPIW) at $0.077$. In particular the PICP captures the same notion as the $ACC_{CI_\alpha}$ introduced above, but computed on the predictions intervals instead of the true values intervals.

This approach to build confidence intervals for performance predictions is very simple yet producing intervals with good coverage and can serve as a robust baseline for future works on uncertainty estimation of performance predictors.

\section{Experiments}

\label{sec:experiments}

Our experimental protocol focuses on evaluating the ability of performance predictors to generalize to unseen shifts, representing a real-world situation where we deploy a model in an unknown domain. To this purpose, we investigate the following scenarios;
\begin{itemize}
\item when a shift seen at test-time has been encountered at train-time with a different severity (unseen\_severity),
\item when the shift seen at test-time is a synthetic perturbation, or a subpopulation shift, that has not been encountered at train-time (unseen\_shift and unseen\_subpop\_shift respectively),
\item when the shift results from a real-world domain change (natural).
\item when we predict the performance on data that follows the same distribution as the source data (no\_shift). It is a controlled situation for which we expect to predict a null performance drop.
\end{itemize}

In the following we give an overview of the experiments, while details are presented in Subsection \ref{subsec:exp_details} of the Appendix. A study of the sensitivity of the error predictor to the number of training shift types and number of training domains is also reported in the Subsection \ref{subsec:exp_incr} of the Appendix.

\subsection{Datasets and Primary Task}

We evaluate the performance predictors on ten real-world datasets for classification listed in Table \ref{tab:datasets}. For all datasets, the \textit{primary task} is a classification task with a Random Forest Classifier as \textit{primary model}, with default hyper-parameters and calibrated via Platt's scaling, trained on a source (train) dataset. We report the average accuracy across $10$ runs in Table \ref{tab:datasets}. We compute the performance drop as the (signed) difference of the primary model accuracy on a target dataset and its accuracy on the source (test) set. For the sake of clarity, we highlight the results obtained on the Video-Games dataset since they are consistent with the outcomes obtained for all the other datasets, as presented in the Appendix.




\begin{table*}[t]
\caption{Real-world classification datasets}
\label{tab:datasets}
\begin{adjustbox}{width=\textwidth,center}
\begin{tabular}{lcccllc}
\toprule
Dataset & classes & features & size & task & source split & RF accuracy\\
\midrule
Adult                     & 2 & 14 & 48842 & low income/high income & race:White & 0.826 \tiny{$\pm$0.018}\\
Artificial characters &  10 & 7 & 10218 & char class & V1:0 & 0.566 \tiny{$\pm$0.026}  \\
Bank                     & 2 & 16 & 11162 & default & marital:married & 0.983 \tiny{$\pm$0.004}\\
Bng Zoo               & 8  & 17 & 1M & animal type &  catsize:false & 0.917 \tiny{$\pm$0.014} \\
Bng Ionosphere   & 2 & 34 & 1M & ionosphere class  & a20:B2of3 & 0.931 \tiny{$\pm$0.012}  \\
Default of credit card clients & 2 & 24 & 30000 & default &  SEX:2 & 0.829 \tiny{$\pm$0.007} \\
Heart                    & 2 & 12 & 70000 &  cardiovascular disease &  gender:1 & 0.708 \tiny{$\pm$0.014}\\
JSBach chorals    & 7 & 16 & 5665 & chord label & meter:3 &  0.832 \tiny{$\pm$0.012}  \\
SDSS                   & 3 & 17 & 10000 & space object &  camcol:4 & 0.986 \tiny{$\pm$0.004}  \\
Video-Games       & 2 & 11 & 14073 &  low sales/high sales & genre:Action & 0.796 \tiny{$\pm$0.019}\\
\bottomrule
\end{tabular}
\end{adjustbox}
\end{table*}

\subsection{Distribution Shifts}
\begin{figure}[b!]
\begin{center}
\centerline{\includegraphics[width=\columnwidth]{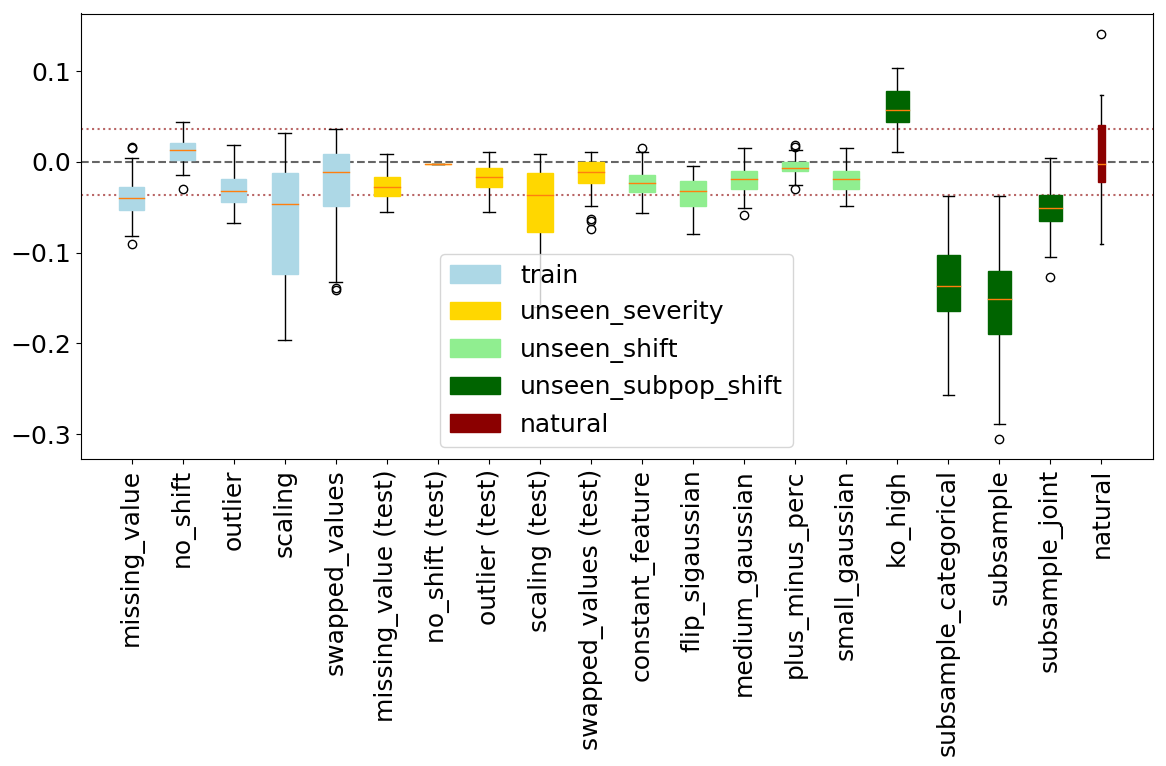}}
\caption{Synthetic and natural shifts true accuracy drop for the Video-Games dataset. The red lines show the confidence interval at 95\% significance level for the accuracy in i.i.d. setting.}
\label{fig:real_true_drop}
\end{center}
\end{figure}

We detail how we build both \textit{synthetic} and \textit{natural} shifts. The impact of the various shift types is visible in Figure \ref{fig:real_true_drop} for the Video-Games dataset.
\subsubsection{Synthetic Shifts}
\label{subsec:shifts}
In Table \ref{tab:shifts}, we describe the various types of synthetic shifts applied to datasets. The implementation of these corruptions is publicly available\footnote{\url{https://github.com/dataiku-research/drift_dac}} and are derived from \cite{Lipton18} and \cite{Redyuk2019}. We apply the same set of perturbations in the training and test shifts, a different set of perturbations in the unseen\_shift and unseen\_subpop\_shift sets, as detailed in Table \ref{tab:shifts}.  The parameters controlling the shift severity are the fraction of samples and fraction of features under shift \cite{Maggio21}. The training set and test shift differ by applying a different severity.
\begin{table*}[h]
\caption{Simulated shift types.}
\label{tab:shifts}
\begin{adjustbox}{width=\textwidth,center}
\begin{tabular}{lll}
\toprule
Set & Shift type & Description \\
\midrule
\multirow{4}{*}{Train/Test Unseen Severity} & SwappedValues & swaps the values of two randomly selected features on $s$\% of samples and $f$\% of features. \\
\cmidrule{2-3}
 & Scaling & add a random fixed value to the features on $s$\% of samples and $f$\% of features. \\
 \cmidrule{2-3}
 & Outliers & Gaussian noise addition with standard deviation different and randomly sampled for each shifted feature on $s$\% of samples and $f$\% of features. \\
 \cmidrule{2-3}
 & MissingValues & replace the value of a fraction of samples with a missing value indicator on $s$\% of samples  and $f$\% of features. \\
\midrule
\multirow{5}{*}{Test Unseen Shift} & Small Gaussian & Small amount of Gaussian Noise applied on $s$\% of samples and $f$\% of features. \\
\cmidrule{2-3}
 & Medium Gaussian & Medium amount of Gaussian Noise applied on $s$\% of samples and $f$\% of features. \\
 \cmidrule{2-3}
 & FlipSign & randomly flips the sign of numeric features on $s$\% of samples and $f$\% of features. \\
 \cmidrule{2-3}
 & ConstantNumeric & assigns a random constant value to a fraction of samples on $s$\% of samples and $f$\% of features. \\
 \cmidrule{2-3}
 & PlusMinusSomePercent & add the feature value its $p\%$ for on $s$\% of samples and $f$\% of features. \\
\midrule
\multirow{4}{*}{Test Unseen Subpop Shift} & Joint Subsampling & Keeps an observation with probability decreasing as points are  away from the samples mean. \\
 \cmidrule{2-3}
    & SubsamplingNumeric & Subsample with low probability samples with low feature values separately for $f$\% numeric  features.  \\
  \cmidrule{2-3}
     & SubsamplingCategorical & Subsample with low probability samples in a random range of categories separately for $f$\% categorical features.  \\
  \cmidrule{2-3}
      & Knock-out & Remove $s$\% of majority class. \\
\bottomrule
\end{tabular}
\end{adjustbox}
\end{table*}

\subsubsection{Natural Shifts}
Following \cite{Upadhyay2021}, we assume that a \textit{natural} domain change occurs when splitting a dataset for different values of a chosen feature, named a split variable. After splitting, we obtain the source and target domains by removing the split variable of the datasets. We describe the split variable used for the different datasets in Table \ref{tab:datasets}.



\subsection{Robustness to Unseen Shifts}
\label{subsec:exp_robustness}
\begin{figure}[h!]
\begin{center}
\centerline{\includegraphics[width=\columnwidth]{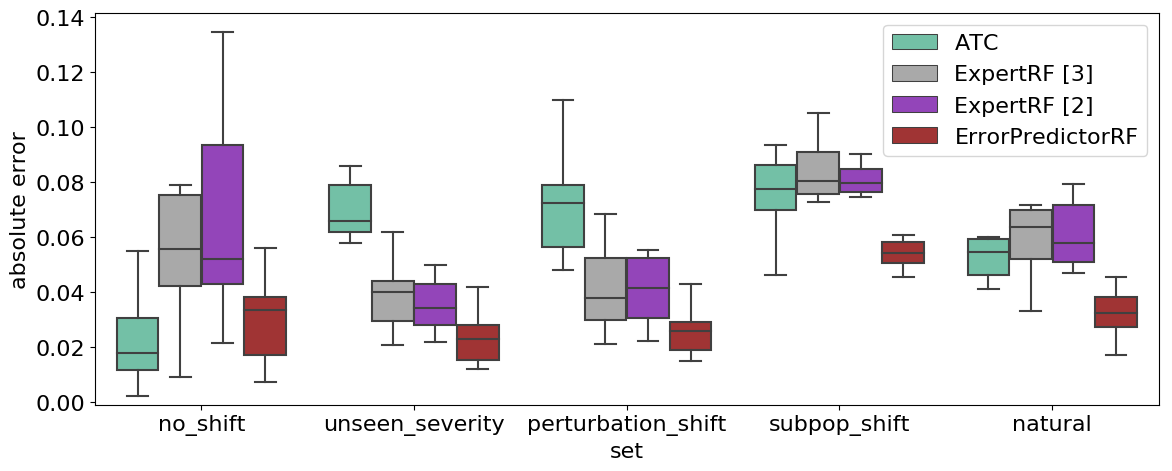}}
\caption{Absolute drop prediction error on Video Games for different test sets.}
\label{fig:robustness_video_games}
\end{center}
\end{figure}

\begin{figure}[h!]
\begin{center}
\centerline{\includegraphics[width=\columnwidth]{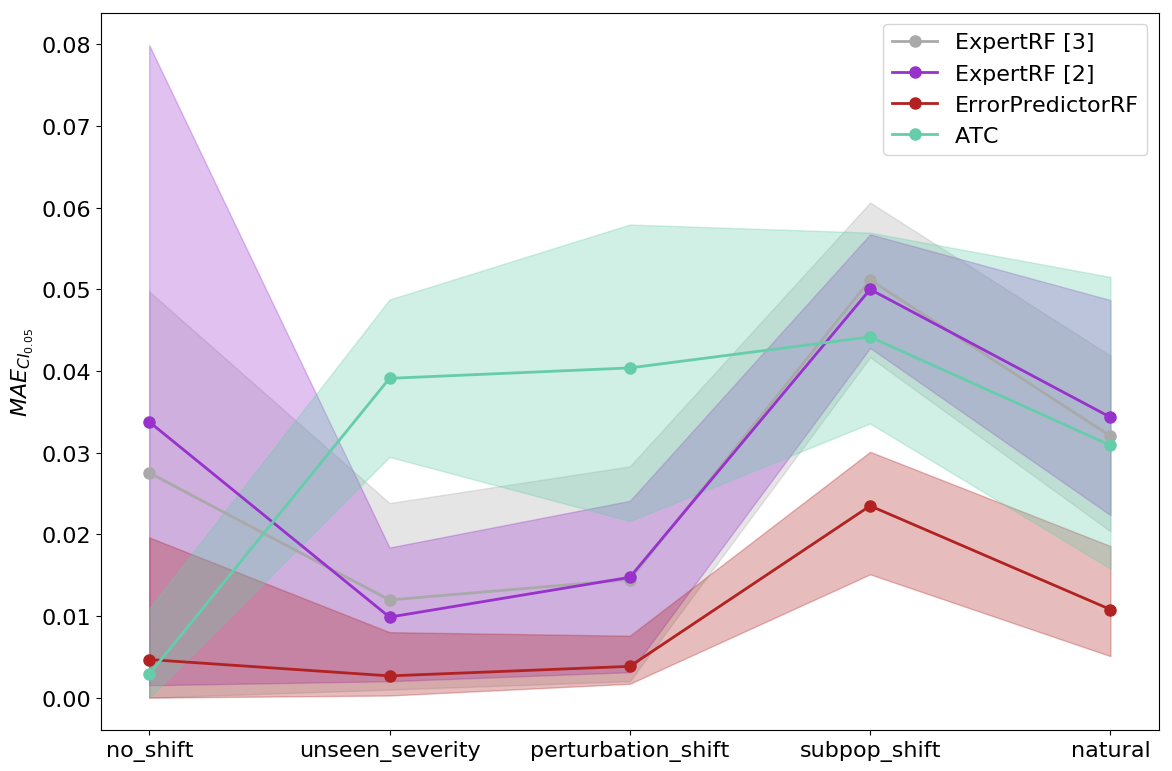}}
\caption{$MAE_{CI_{0.05}}$ metric on Video Games for different test sets.}
\label{fig:mae_ci_video_games}
\end{center}
\end{figure}

In this experiment we compare the ability of performance predictors to generalize to unseen severities, shift types and natural domain changes. Figure \ref{fig:robustness_video_games} 
shows the absolute error of the predictions for the different test sets and highlight that the \textit{Error Predictor} is more robust than other approaches to unseen domain changes. The ATC is generally worse than expert models and systematically under performs with the respect to the error predictor in the considered drift scenarios, except for non drift situations and some natural shifts which are in-distribution. Even taking into account the uncertainty in the measure of the ground truth accuracies due to the finite sample size, we observe that the \textit{Error Predictor} outperforms the other approaches (Figure \ref{fig:mae_ci_video_games}). The metrics for all considered datasets are shown in Tables \ref{tab:unseen_shift_metrics_paper} and  \ref{tab:unseen_natural_metrics_paper} (and Tables \ref{tab:unseen_no_shift_metrics_paper} to  \ref{tab:unseen_subpop_shift_metrics_paper} of the Appendix) for various shift scenarios and confirm what illustrated above for the Video Games dataset: the error predictor trained on synthetically shifted data outperforms the expert models in all cases and it also beats the ATC strategy, especially when dealing with unseen synthetic shifts.



\begin{table*}[h!]
\caption{Performance Predictors ability to generalize to unseen synthetic shifts ($MAE_{CI_{0.05}}$ metric)}
\label{tab:unseen_shift_metrics_paper}
\begin{adjustbox}{width=0.7\textwidth,center}
\begin{tabular}{lcccc}
\toprule
dataset &                       ATC &         ExpertRF \cite{Redyuk2019} &          ExpertRF \cite{Elsahar2019} &          ErrorPredictorRF \\
\midrule
adult                          &  0.025 \tiny{$\pm$ 0.003} &  0.022 \tiny{$\pm$ 0.019} &  0.016 \tiny{$\pm$ 0.011} &  \textbf{0.006} \tiny{$\pm$ 0.005} \\
artificial\_characters          &  0.029 \tiny{$\pm$ 0.006} &  0.076 \tiny{$\pm$ 0.012} &  0.077 \tiny{$\pm$ 0.020} &  \textbf{0.022} \tiny{$\pm$ 0.009} \\
bank                           &  0.040 \tiny{$\pm$ 0.015} &  0.003 \tiny{$\pm$ 0.005} &  0.002 \tiny{$\pm$ 0.003} &  \textbf{0.001} \tiny{$\pm$ 0.001} \\
bng\_ionosphere                 &  0.002 \tiny{$\pm$ 0.004} &  0.300 \tiny{$\pm$ 0.151} &  0.129 \tiny{$\pm$ 0.098} &  \textbf{0.000} \tiny{$\pm$ 0.000} \\
bng\_zoo                        &  \textbf{0.001} \tiny{$\pm$ 0.003} &  0.175 \tiny{$\pm$ 0.068} &  0.123 \tiny{$\pm$ 0.041} &  \textbf{0.001} \tiny{$\pm$ 0.002} \\
default\_of\_credit\_card\_clients &  0.101 \tiny{$\pm$ 0.012} &  0.042 \tiny{$\pm$ 0.026} &  0.035 \tiny{$\pm$ 0.022} &  \textbf{0.020} \tiny{$\pm$ 0.007} \\
heart                          &  0.058 \tiny{$\pm$ 0.012} &  0.044 \tiny{$\pm$ 0.015} &  0.046 \tiny{$\pm$ 0.017} &  \textbf{0.011} \tiny{$\pm$ 0.005} \\
jsbach\_chorals        &  0.004 \tiny{$\pm$ 0.007} &  0.234 \tiny{$\pm$ 0.099} &  0.104 \tiny{$\pm$ 0.064} &  \textbf{0.000} \tiny{$\pm$ 0.000} \\
SDSS                           &  0.114 \tiny{$\pm$ 0.012} &  0.132 \tiny{$\pm$ 0.023} &  0.166 \tiny{$\pm$ 0.030} &  \textbf{0.078} \tiny{$\pm$ 0.007} \\
video\_games                    &  0.040 \tiny{$\pm$ 0.016} &  0.013 \tiny{$\pm$ 0.009} &  0.014 \tiny{$\pm$ 0.010} & \textbf{0.003} \tiny{$\pm$ 0.002} \\
\bottomrule
\end{tabular}
\end{adjustbox}
\end{table*}

\begin{table*}[h!]
\caption{Performance Predictors ability to generalize to unseen natural shifts ($MAE_{CI_{0.05}}$ metric)}
\label{tab:unseen_natural_metrics_paper}
\begin{adjustbox}{width=0.7\textwidth,center}
\begin{tabular}{lcccc}
\toprule
dataset &                       ATC &         ExpertRF \cite{Redyuk2019} &          ExpertRF \cite{Elsahar2019} &          ErrorPredictorRF \\
\midrule
adult                          &  0.004 \tiny{$\pm$ 0.006} &  0.071 \tiny{$\pm$ 0.034} &  0.063 \tiny{$\pm$ 0.025} &  \textbf{0.001} \tiny{$\pm$ 0.002} \\
artificial\_characters          &  \textbf{0.083} \tiny{$\pm$ 0.042} &  0.194 \tiny{$\pm$ 0.031} &  0.204 \tiny{$\pm$ 0.056} &  0.217 \tiny{$\pm$ 0.037} \\
bank                           &  \textbf{0.001} \tiny{$\pm$ 0.001} &  0.000 \tiny{$\pm$ 0.001} &  0.000 \tiny{$\pm$ 0.000} &  \textbf{0.001} \tiny{$\pm$ 0.001} \\
bng\_ionosphere                 &  0.014 \tiny{$\pm$ 0.016} &  0.145 \tiny{$\pm$ 0.163} &  0.093 \tiny{$\pm$ 0.060} &  \textbf{0.011} \tiny{$\pm$ 0.007} \\
bng\_zoo                        &  \textbf{0.000} \tiny{$\pm$ 0.000} &  0.459 \tiny{$\pm$ 0.111} &  0.201 \tiny{$\pm$ 0.109} &  \textbf{0.000} \tiny{$\pm$ 0.000} \\
default\_of\_credit\_card\_clients &  \textbf{0.001} \tiny{$\pm$ 0.002} &  0.020 \tiny{$\pm$ 0.036} &  0.021 \tiny{$\pm$ 0.027} &  0.020 \tiny{$\pm$ 0.012} \\
heart                          &  0.008 \tiny{$\pm$ 0.018} &  0.046 \tiny{$\pm$ 0.023} &  0.058 \tiny{$\pm$ 0.042} &  \textbf{0.005} \tiny{$\pm$ 0.009} \\
jsbach\_chorals        &  \textbf{0.008} \tiny{$\pm$ 0.007} &  0.326 \tiny{$\pm$ 0.037} &  0.152 \tiny{$\pm$ 0.095} &  0.020 \tiny{$\pm$ 0.009} \\
SDSS                           &  \textbf{0.002} \tiny{$\pm$ 0.003} &  0.199 \tiny{$\pm$ 0.053} &  0.166 \tiny{$\pm$ 0.070} &  0.013 \tiny{$\pm$ 0.012} \\
video\_games                    &  0.031 \tiny{$\pm$ 0.016} &  0.032 \tiny{$\pm$ 0.009} &  0.035 \tiny{$\pm$ 0.012} &  \textbf{0.011} \tiny{$\pm$ 0.005} \\
\bottomrule
\end{tabular}
\end{adjustbox}
\end{table*}

\subsection{Impact of Data Augmentation}

\begin{figure}[t]
\begin{center}
\centerline{\includegraphics[width=\columnwidth]{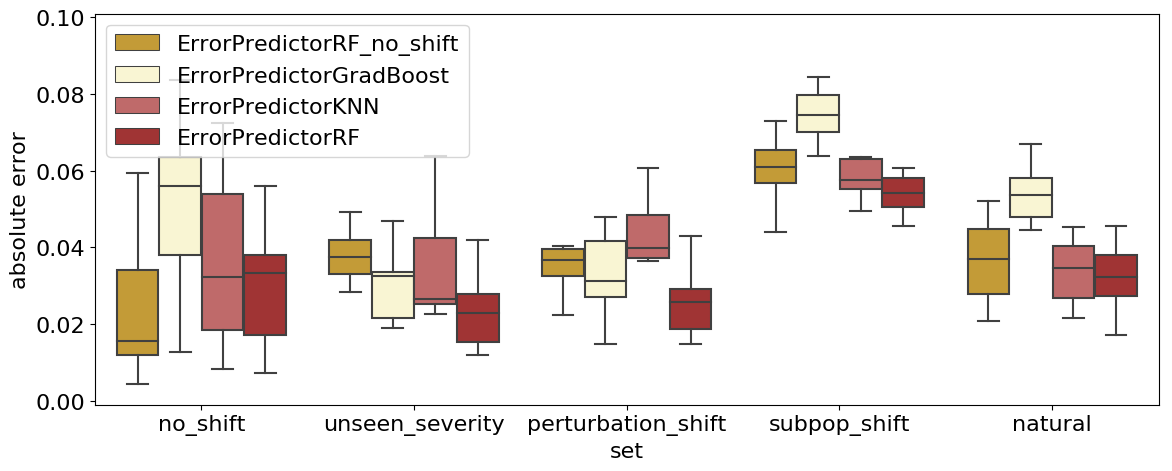}}
\caption{Absolute drop prediction error on Video Games for different variants of the \textit{Error Predictor}.}
\label{fig:errpred_video_games}
\end{center}
\end{figure}

In Figure
\ref{fig:errpred_video_games} we can observe an important increase in the generalization ability of the \textit{Error Predictor} when trained on augmented synthetic shifted datasets, with respect to the state-of-the-art approach of training the \textit{Error Predictor} on the clean test set \cite{Elder20}. 
The prediction error is significantly reduced ond real-world dataset when data augmentation is employed (\textit{ErrorPredictorRF} vs. \textit{ErrorPredictorRF\_no\_shift} in the figure).


\subsection{Impact of the Error Model}

The typical choice for the binary classifier underlying the error predictor is the GradientBoosting classifier \cite{Elder20}, but we observe that this choice is crucial for the generalization ability of the error predictor.
In Figure
\ref{fig:errpred_video_games} we compare three choices for the underlying model in the error predictor: GradientBoosting, kNN, and RandomForest. In spite of being the default choice for the error model, the GradientBoosting is far from optimal in terms of generalization abilities to unseen shifts. The KNN can sometimes be comparable to the Random Forest, but the latter outperforms other choices in all experiments on real-world datasets.

\section{Conclusions}
\label{sec:conclusion}

Predicting Out-Of-Distribution performance is crucial to monitor ML models deployed in production. In this work we compare the latest approaches for performance predictions and evaluate their ability to generalize to new shifts encountered in production. We perform a benchmark on ten classification datasets, proposing a protocol to simulate dataset shifts and to derive natural shifts for evaluation, and show that drift features used in prior works are not expressive enough to generalize under distribution shift. We show experimentally that the approach aggregating the sample-wise error probability is more robust than approaches based on expert features and on thresholding the model confidence.
Our results also highlight that training data augmentation based on simulated drift scenarios improves significantly the generalization ability of state-of-the-art performance predictors. In addition we present a natural and effortless uncertainty estimation of the predicted accuracy for reliable use of any performance predictor under distribution shift. 

\bibliography{main}

\begin{thebibliography}{10}
\providecommand{\url}[1]{#1}
\csname url@samestyle\endcsname
\providecommand{\newblock}{\relax}
\providecommand{\bibinfo}[2]{#2}
\providecommand{\BIBentrySTDinterwordspacing}{\spaceskip=0pt\relax}
\providecommand{\BIBentryALTinterwordstretchfactor}{4}
\providecommand{\BIBentryALTinterwordspacing}{\spaceskip=\fontdimen2\font plus
\BIBentryALTinterwordstretchfactor\fontdimen3\font minus
  \fontdimen4\font\relax}
\providecommand{\BIBforeignlanguage}[2]{{%
\expandafter\ifx\csname l@#1\endcsname\relax
\typeout{** WARNING: IEEEtran.bst: No hyphenation pattern has been}%
\typeout{** loaded for the language `#1'. Using the pattern for}%
\typeout{** the default language instead.}%
\else
\language=\csname l@#1\endcsname
\fi
#2}}
\providecommand{\BIBdecl}{\relax}
\BIBdecl

\bibitem{Elder20}
B.~{Elder} \emph{et~al.}, ``{Learning Prediction Intervals for Model
  Performance},'' \emph{Proceedings of the AAAI Conference on Artificial
  Intelligence}, 2020.

\bibitem{Elsahar2019}
H.~Elsahar and M.~Gall{\'{e}}, ``{To Annotate or Not? Predicting Performance
  drop under domain shift},'' 2019.

\bibitem{Redyuk2019}
S.~Redyuk \emph{et~al.}, ``{Learning to validate the predictions of black box
  machine learning models on unseen data},'' \emph{Proceedings of the ACM
  SIGMOD International Conference on Management of Data}, 2019.

\bibitem{Garg2022}
S.~{Garg} \emph{et~al.}, ``{Leveraging Unlabeled Data to Predict
  Out-of-Distribution Performance},'' \emph{Proceedings of the International
  Conference on Learning Representations, ICLR}, 2022.

\bibitem{Ovadia2019}
Y.~{Ovadia} \emph{et~al.}, ``{{Can You Trust Your Model's Uncertainty?
  Evaluating Predictive Uncertainty Under Dataset Shift}},'' \emph{arXiv
  e-prints}, p. arXiv:1906.02530, Jun. 2019.

\bibitem{DAmour2020}
A.~D'Amour \emph{et~al.}, ``{Underspecification Presents Challenges for
  Credibility in Modern Machine Learning}.''

\bibitem{Chen2019}
T.~Chen \emph{et~al.}, ``Confidence scoring using whitebox meta-models with
  linear classifier probes,'' \emph{22nd International Conference on Artificial
  Intelligence and Statistics}, pp. 1467--1475, 2008.

\bibitem{Guerra2008}
S.~B. Guerra \emph{et~al.}, ``Predicting the performance of learning algorithms
  using support vector machines as meta-regressors,'' \emph{International
  Conference on Artificial Neural Networks}, vol. 5163, 2008.

\bibitem{Schat2020}
\BIBentryALTinterwordspacing
E.~Schat \emph{et~al.}, ``The data representativeness criterion: Predicting the
  performance of supervised classification based on data set similarity,''
  \emph{PLOS ONE}, vol.~15, no.~8, p. e0237009, Aug 2020. [Online]. Available:
  \url{http://dx.doi.org/10.1371/journal.pone.0237009}
\BIBentrySTDinterwordspacing

\bibitem{Talagala2019}
T.~S. Talagala \emph{et~al.}, ``Fformpp: Feature-based forecast model
  performance prediction,'' 2021.

\bibitem{Deng2021-2}
W.~Deng and L.~Zheng, ``Are labels always necessary for classifier accuracy
  evaluation?'' \emph{Proceedings ofthe IEEE/CVF Conference on Computer Vision
  and Pattern Recognition}, p. 15069–15078, 2021.

\bibitem{Deng2021}
W.~Deng \emph{et~al.}, ``What does rotation prediction tell us about classifier
  accuracy under varying testing environments?'' 2021.

\bibitem{Guillory2021}
D.~Guillory \emph{et~al.}, ``Predicting with confidence on unseen
  distributions,'' 2021.

\bibitem{Taori2020}
R.~Taori \emph{et~al.}, ``{Measuring Robustness to Natural Distribution Shifts
  in Image Classification},'' \emph{arXiv}, 2020.

\bibitem{Ben-david2010}
S.~Ben-David \emph{et~al.}, ``{A theory of learning from different domains},''
  \emph{Mach Learn}, vol.~79, p. 151–175, 2010.

\bibitem{Pishro2014}
H.~Pishro-Nik, \emph{Introduction to probability, statistics, and random
  processes}.\hskip 1em plus 0.5em minus 0.4em\relax Kappa Research LLC, 2014.

\bibitem{Lipton18}
Z.~C. {Lipton} \emph{et~al.}, ``{{Detecting and Correcting for Label Shift with
  Black Box Predictors}},'' \emph{arXiv e-prints}, p. arXiv:1802.03916, Feb.
  2018.

\bibitem{Maggio21}
\BIBentryALTinterwordspacing
S.~Maggio and L.~Dreyfus{-}Schmidt, ``Ensembling shift detectors: an extensive
  empirical evaluation,'' \emph{ECML PKDD}, vol. abs/2106.14608, 2021.
  [Online]. Available: \url{https://arxiv.org/abs/2106.14608}
\BIBentrySTDinterwordspacing

\bibitem{Upadhyay2021}
\BIBentryALTinterwordspacing
S.~Upadhyay \emph{et~al.}, ``{Towards Robust and Reliable Algorithmic
  Recourse},'' 2021. [Online]. Available: \url{http://arxiv.org/abs/2102.13620}
\BIBentrySTDinterwordspacing

\end{thebibliography}
\bibliographystyle{IEEEtran}

\newpage
\appendix

\section{Appendix}

\subsection{Experimental Details and Reproducibility}
\label{subsec:exp_details}

The implementation of dataset shifts and the experimental code are available in this github repository \footnote{\url{https://github.com/dataiku-research/performance\_prediction\_under\_shift}}.




The performance predictors are all trained on a fixed number of synthetic shift types with various high severity of shift (proportion of samples affected) on various proportions of features (Table \ref{tab:shifts}).

Training performance drop predictors requires a meta-learning framework, where one single training observation is an entire dataset, under a given drift type and with a given performance drop.
Each dataset is first split with respect to the split domain variable in order to retain one source domain only (i.e. Genre:Action, Race:White, ...) and use the others to build the natural target datasets. 

The number of samples $n_{samples}$ in a dataset built for the experiments depends on the size of the complete dataset, i.e. $500$ for Video Games and Adult. We run each experiment $10$ times with different random seeds.

The source domain is sampled in:
\begin{itemize}
\item primary training set: $n_{samples}$ samples representing the clean source dataset.
\item $2 \cdot n_{samples}$ samples constituting the source pool. At each new generation of a shift scenario, we draw $n_{samples}$ samples from this pool to build the initial clean dataset to which the shift is applied. This sampling allows to collect more diverse shift scenarios for the training (and depart from the protocol proposed in \cite{Redyuk2019} where one single source dataset is used).
\item primary validation set: $n_{samples}$ clean samples representing a reference dataset used to score the primary model and with respect to which to compute the drift metrics. This set is sampled from the previous source pool.
\item primary target set: separate $n_{samples}$ samples representing the clean target dataset, to which shifts are applied to build all test datasets (unseen\_severity, unseen\_shift, unseen\_subpop\_shift, natural).
\end{itemize}

For all shifts the proportion of features affected is randomly sampled in $[0.25, 0.95]$.

For the training data we generate $100$ drift scenarios with severity randomly sampled in $[0.75, 0.95]$ for each drift type, thus collecting $5 \cdot 100$ training datasets overall.

Likewise, for the unseen\_severity we generate $100$ drift scenarios for each of the same synthetic shifts as before, but with various severities, randomly sampled in a range of $[0.25, 0.74]$. Thus this set also contains $5 \cdot 100$ datasets.

Similarly for the unseen\_shift and the unseen\_subpop\_shift: we generate $100$ drift scenarios for each other different synthetic shifts with various severities, randomly sampled in a range of $[0.25, 0.95]$.

For all the meta-sets of datasets above, each shift is applied to a different clean dataset, sampled from the source pool, in order to ensure diversity in the generated drift scenarios.

In order to build the natural drift scenarios, natural, we draw $n_{samples}$ samples for any different value of the split variable other than the one selected for the source dataset. Depending on the categories of the split variable of the dataset, we have 10 natural drift scenarios for Video Games and 2 for Adult.

The above process is realized for 1 single source domain (i.e. \textit{Genre}:\textit{Action};  \textit{Race}: \textit{White}, ...).

\subsection{Impact of n. of Training Shift Types}
\label{subsec:exp_incr}

In the previous experiment, we fixed the number of synthetic shift types to $5$. In this section we study the impact of the number of training shift types and the number of domains per shift type on the performance of the \textit{Error Predictor}.
 
\begin{figure*}[t!]
\centering
\subfloat[Sensitivity of \textit{Error Predictor} to number of training shift type]{
\includegraphics[width=0.45\linewidth]{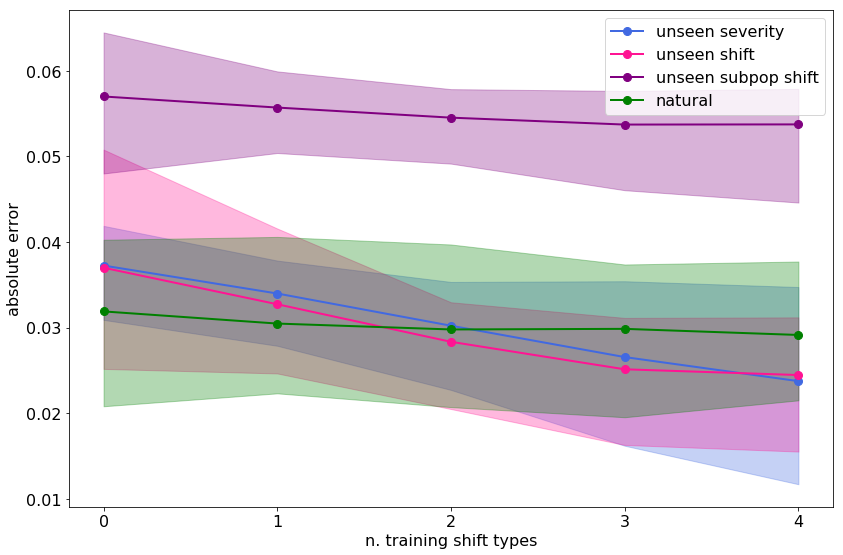}
\label{fig:sensitivity_to_n_shifts}
}
\subfloat[Sensitivity of \textit{Error Predictor} to number of training domains per shift type]{
\includegraphics[width=0.45\linewidth]{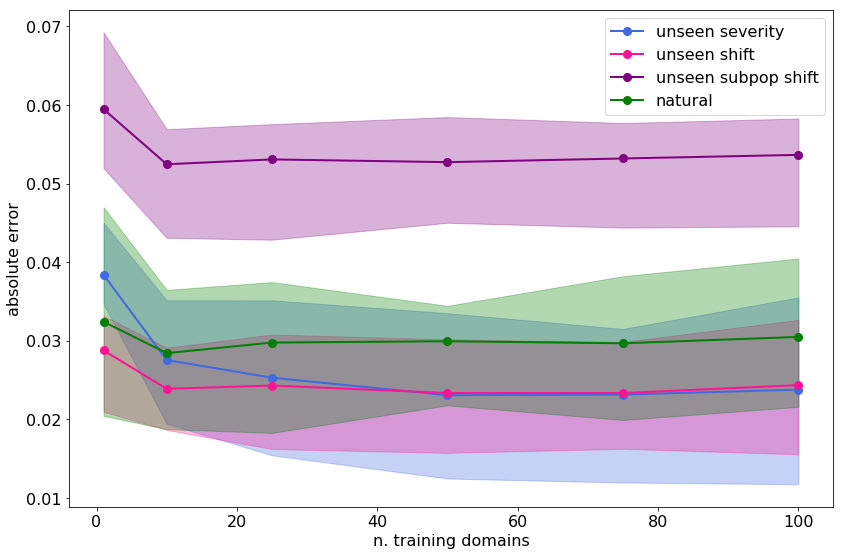}
\label{fig:sensitivity_to_n_domains}
}
\caption{Absolute drop prediction error, varying the number of synthetic shift types and the number of shifted datasets per shift type used during training.}
\end{figure*}

Figure \ref{fig:sensitivity_to_n_shifts} shows that on the Video Games dataset increasing the number of training shift types is beneficial especially for perturbation shifts while it only slightly impacts performances on the natural and subpopulation shifts. We report results on 10 runs for this experiment, where the variance is also due to the different impact given by different shift types, as we consider for each $k$ of the x-axis all possible combinations of $k$ shift types among the $5$ available for the training.

Given a fixed number of training shifts, we also study the optimal number of shifted datasets per each shift type should be used in the training, or equivalently the number of levels of severity per each shift type. Figure \ref{fig:sensitivity_to_n_domains} shows that using more than 10 datasets per shift type is not beneficial.  

\subsection{Results on all datasets and drift scenarios}

In addition to Tables \ref{tab:unseen_shift_metrics_paper} and  \ref{tab:unseen_natural_metrics_paper} of Subsection \ref{subsec:exp_robustness}, we report here the results for the non shift and subpopulation shifts scenarios, where ATC and ErrorPredictorRF perform similarly, as well as results for the unseen severity scenarios where the ErrorPredictorRF is a better approach.

\begin{table*}[h!]
\caption{Performance Predictors ability to generalize to unseen non shift scenarios ($MAE_{CI_{0.05}}$ metric)}
\label{tab:unseen_no_shift_metrics_paper}
\begin{adjustbox}{width=0.7\textwidth,center}
\begin{tabular}{lcccc}
\toprule
dataset  &                       ATC &         ExpertRF \cite{Redyuk2019} &          ExpertRF \cite{Elsahar2019} &          ErrorPredictorRF \\
\midrule
adult                          &  \textbf{0.000} \tiny{$\pm$ 0.001} &  0.023 \tiny{$\pm$ 0.031} &  0.018 \tiny{$\pm$ 0.023} &  0.001 \tiny{$\pm$ 0.003} \\
artificial\_characters          &  \textbf{0.002} \tiny{$\pm$ 0.006} &  0.138 \tiny{$\pm$ 0.034} &  0.158 \tiny{$\pm$ 0.048} &  0.031 \tiny{$\pm$ 0.023} \\
bank                           &  0.001 \tiny{$\pm$ 0.002} &  0.000 \tiny{$\pm$ 0.000} &  0.000 \tiny{$\pm$ 0.001} &  \textbf{0.000} \tiny{$\pm$ 0.000} \\
bng\_ionosphere                 &  0.002 \tiny{$\pm$ 0.004} &  0.300 \tiny{$\pm$ 0.151} &  0.132 \tiny{$\pm$ 0.102} &  \textbf{0.000} \tiny{$\pm$ 0.000} \\
bng\_zoo                        &  \textbf{0.001} \tiny{$\pm$ 0.003} &  0.179 \tiny{$\pm$ 0.148} &  0.099 \tiny{$\pm$ 0.059} &  \textbf{0.001} \tiny{$\pm$ 0.003} \\
default\_of\_credit\_card\_clients &  \textbf{0.006} \tiny{$\pm$ 0.007} &  0.035 \tiny{$\pm$ 0.053} &  0.036 \tiny{$\pm$ 0.043} &  0.010 \tiny{$\pm$ 0.007} \\
heart                          &  0.004 \tiny{$\pm$ 0.005} &  0.068 \tiny{$\pm$ 0.021} &  0.072 \tiny{$\pm$ 0.038} &  \textbf{0.003} \tiny{$\pm$ 0.006} \\
jsbach\_chorals        &  0.003 \tiny{$\pm$ 0.007} &  0.235 \tiny{$\pm$ 0.117} &  0.107 \tiny{$\pm$ 0.096} &  \textbf{0.000} \tiny{$\pm$ 0.000} \\
SDSS                           &  \textbf{0.002} \tiny{$\pm$ 0.005} &  0.177 \tiny{$\pm$ 0.055} &  0.162 \tiny{$\pm$ 0.082} &  0.003 \tiny{$\pm$ 0.004} \\
video\_games                    &  \textbf{0.003} \tiny{$\pm$ 0.006} &  0.027 \tiny{$\pm$ 0.024} &  0.036 \tiny{$\pm$ 0.033} &  0.005 \tiny{$\pm$ 0.007} \\
\bottomrule
\end{tabular}
\end{adjustbox}
\end{table*}

\begin{table*}[h!]
\caption{Performance Predictors ability to generalize to unseen severities ($MAE_{CI_{0.05}}$ metric)}
\label{tab:unseen_severity_metrics_paper}
\begin{adjustbox}{width=0.7\textwidth,center}
\begin{tabular}{lcccc}
\toprule
dataset &                       ATC &         ExpertRF  \cite{Redyuk2019} &          ExpertRF \cite{Elsahar2019} &          ErrorPredictorRF \\
\midrule
adult                          &  0.031 \tiny{$\pm$ 0.010} &  0.013 \tiny{$\pm$ 0.012} &  0.012 \tiny{$\pm$ 0.011} &  \textbf{0.001} \tiny{$\pm$ 0.001} \\
artificial\_characters          &  0.056 \tiny{$\pm$ 0.010} &  0.046 \tiny{$\pm$ 0.009} &  0.051 \tiny{$\pm$ 0.015} &  \textbf{0.010} \tiny{$\pm$ 0.007} \\
bank                           &  0.036 \tiny{$\pm$ 0.010} &  0.001 \tiny{$\pm$ 0.003} &  0.001 \tiny{$\pm$ 0.002} &  \textbf{0.000} \tiny{$\pm$ 0.000} \\
bng\_ionosphere                 &  0.131 \tiny{$\pm$ 0.019} &  0.217 \tiny{$\pm$ 0.078} &  0.132 \tiny{$\pm$ 0.046} & \textbf{0.050} \tiny{$\pm$ 0.007} \\
bng\_zoo                        &  0.062 \tiny{$\pm$ 0.008} &  0.130 \tiny{$\pm$ 0.042} &  0.136 \tiny{$\pm$ 0.025} &  \textbf{0.011} \tiny{$\pm$ 0.003} \\
default\_of\_credit\_card\_clients &  0.132 \tiny{$\pm$ 0.029} &  0.033 \tiny{$\pm$ 0.024} &  0.036 \tiny{$\pm$ 0.033} &  \textbf{0.006} \tiny{$\pm$ 0.002} \\
heart                          &  0.071 \tiny{$\pm$ 0.012} &  0.029 \tiny{$\pm$ 0.008} &  0.032 \tiny{$\pm$ 0.011} &  \textbf{0.004} \tiny{$\pm$ 0.002} \\
jsbach\_chorals        &  0.027 \tiny{$\pm$ 0.014} &  0.118 \tiny{$\pm$ 0.058} &  0.091 \tiny{$\pm$ 0.027} &  \textbf{0.002} \tiny{$\pm$ 0.001} \\
SDSS                           &  0.090 \tiny{$\pm$ 0.011} &  0.108 \tiny{$\pm$ 0.020} &  0.141 \tiny{$\pm$ 0.045} &  \textbf{0.023} \tiny{$\pm$ 0.004} \\
video\_games                    &  0.039 \tiny{$\pm$ 0.009} &  0.010 \tiny{$\pm$ 0.007} &  0.010 \tiny{$\pm$ 0.007} &  \textbf{0.002} \tiny{$\pm$ 0.002} \\
\bottomrule
\end{tabular}
\end{adjustbox}
\end{table*}

\begin{table*}[h!]
\caption{Performance Predictors ability to generalize to unseen subpopulation shifts ($MAE_{CI_{0.05}}$ metric)}
\label{tab:unseen_subpop_shift_metrics_paper}
\begin{adjustbox}{width=0.7\textwidth,center}
\begin{tabular}{lcccc}
\toprule
dataset &                       ATC &         ExpertRF \cite{Redyuk2019} &          ExpertRF \cite{Elsahar2019} &          ErrorPredictorRF \\
\midrule
adult                          &  \textbf{0.030} \tiny{$\pm$ 0.012} &  0.035 \tiny{$\pm$ 0.012} &  0.041 \tiny{$\pm$ 0.011} &  0.034 \tiny{$\pm$ 0.014} \\
artificial\_characters          &  \textbf{0.011} \tiny{$\pm$ 0.007} &  0.142 \tiny{$\pm$ 0.027} &  0.178 \tiny{$\pm$ 0.027} &  0.037 \tiny{$\pm$ 0.021} \\
bank                           &  0.005 \tiny{$\pm$ 0.002} &  0.006 \tiny{$\pm$ 0.003} &  0.005 \tiny{$\pm$ 0.002} &  \textbf{0.002} \tiny{$\pm$ 0.000} \\
bng\_ionosphere                 &  \textbf{0.029} \tiny{$\pm$ 0.011} &  0.275 \tiny{$\pm$ 0.078} &  0.147 \tiny{$\pm$ 0.023} &  0.031 \tiny{$\pm$ 0.003} \\
bng\_zoo                        &  0.354 \tiny{$\pm$ 0.006} &  0.386 \tiny{$\pm$ 0.035} &  0.350 \tiny{$\pm$ 0.021} &  \textbf{0.353} \tiny{$\pm$ 0.006} \\
default\_of\_credit\_card\_clients &  0.039 \tiny{$\pm$ 0.008} &  0.095 \tiny{$\pm$ 0.048} &  0.071 \tiny{$\pm$ 0.027} &  \textbf{0.031} \tiny{$\pm$ 0.003} \\
heart                          &  0.016 \tiny{$\pm$ 0.005} &  0.080 \tiny{$\pm$ 0.014} &  0.079 \tiny{$\pm$ 0.023} &  \textbf{0.014} \tiny{$\pm$ 0.005} \\
jsbach\_chorals        &  \textbf{0.334} \tiny{$\pm$ 0.009} &  0.350 \tiny{$\pm$ 0.025} &  0.321 \tiny{$\pm$ 0.041} &  0.335 \tiny{$\pm$ 0.007} \\
SDSS                           &  \textbf{0.009} \tiny{$\pm$ 0.012} &  0.245 \tiny{$\pm$ 0.037} &  0.201 \tiny{$\pm$ 0.059} &  0.012 \tiny{$\pm$ 0.008} \\
video\_games                    &  0.044 \tiny{$\pm$ 0.012} &  0.048 \tiny{$\pm$ 0.012} &  0.049 \tiny{$\pm$ 0.008} &  \textbf{0.024} \tiny{$\pm$ 0.007} \\
\bottomrule
\end{tabular}
\end{adjustbox}
\end{table*} 

\subsection{Results on all datasets}

\begin{figure*}[h]
\centering
\subfloat[adult]{
  \includegraphics[width=.45\linewidth]{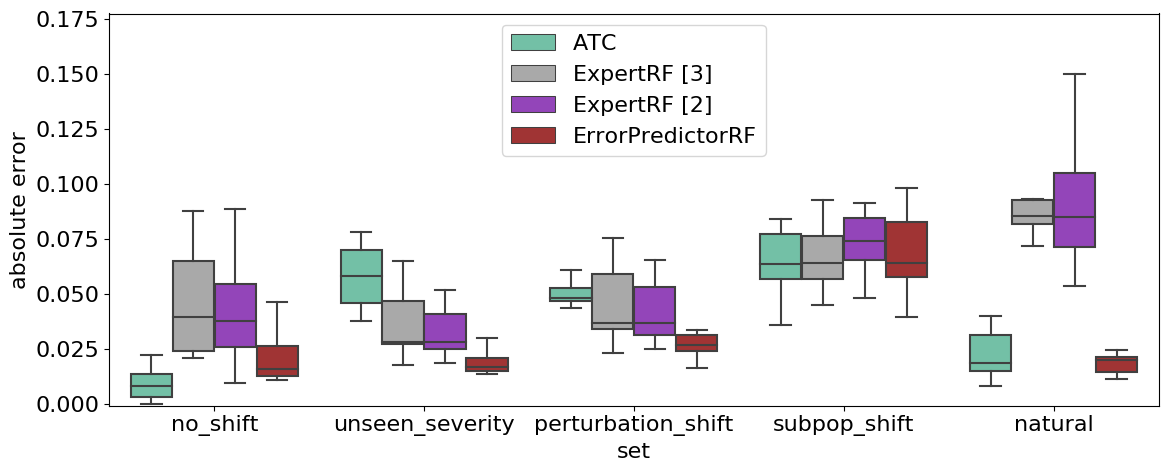}  
}
\subfloat[artificial\_characters]{
  \includegraphics[width=.45\linewidth]{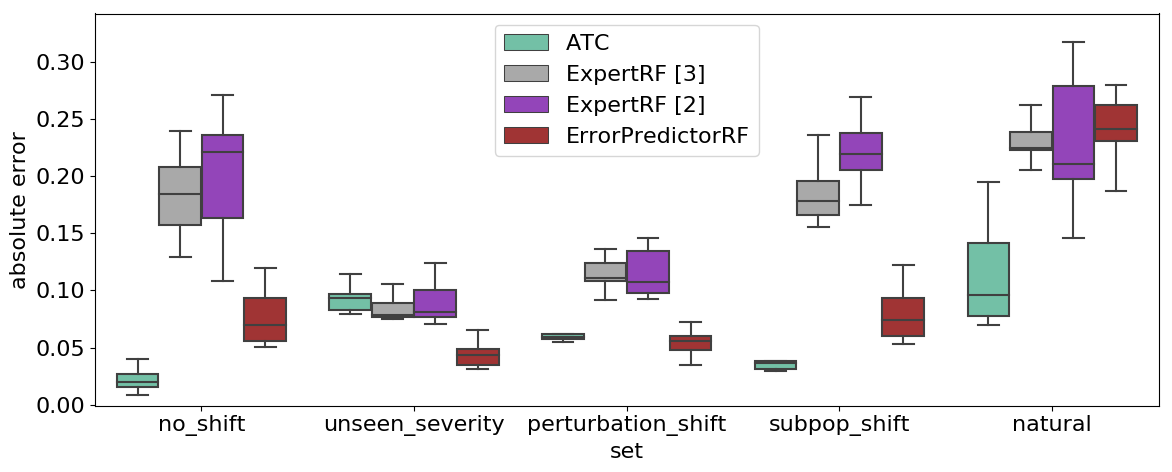}  
}\\
\subfloat[bank]{
  \includegraphics[width=.45\linewidth]{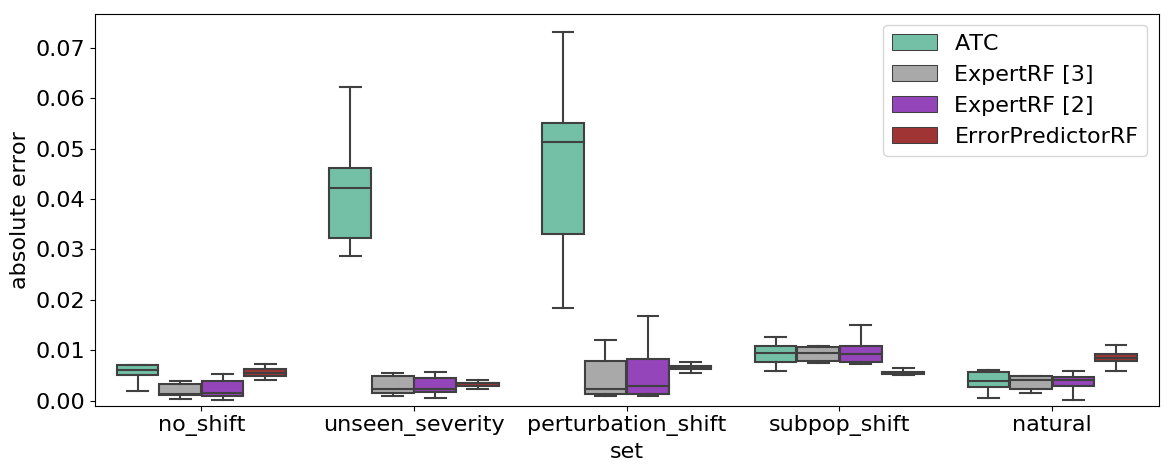}  
}
\subfloat[bng\_ionosphere]{
  \includegraphics[width=.45\linewidth]{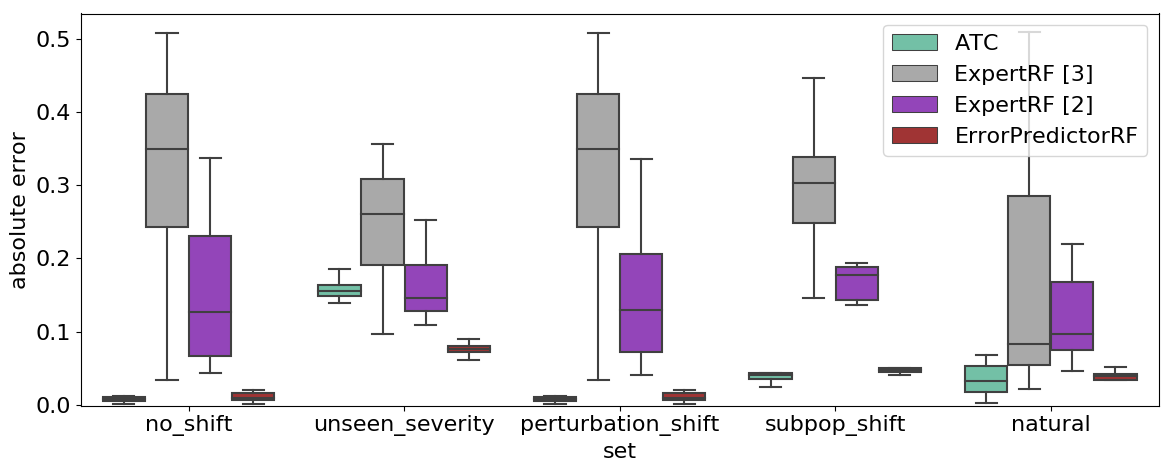}  
}\\
\subfloat[bng\_zoo]{
  \includegraphics[width=.45\linewidth]{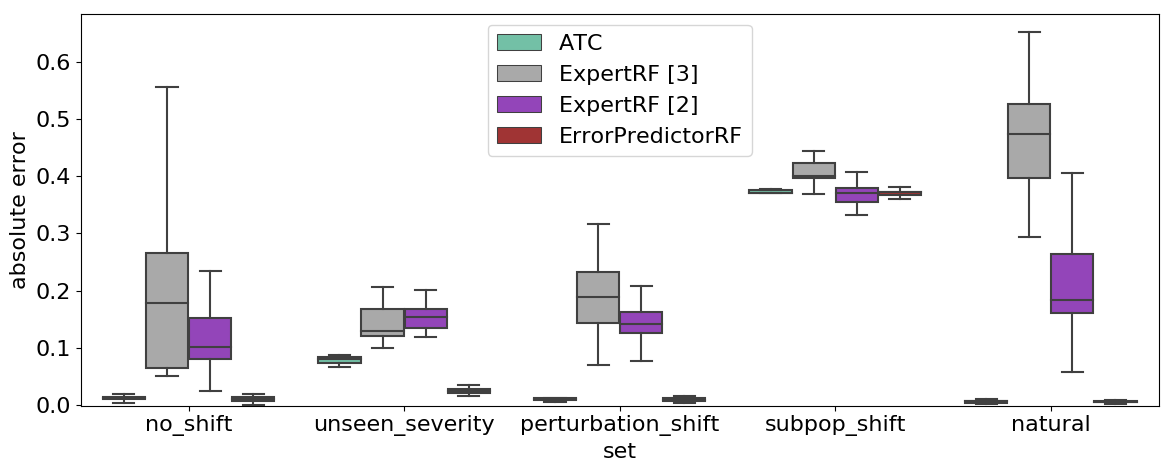}  
}
\subfloat[default\_of\_credit\_card\_clients]{
  \includegraphics[width=.45\linewidth]{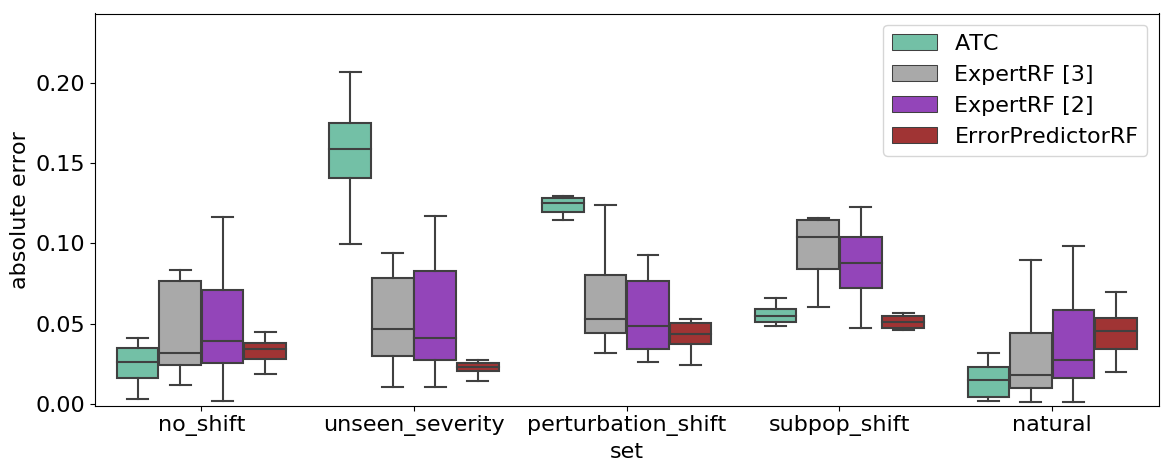}  
}\\
\subfloat[heart]{
  \includegraphics[width=.45\linewidth]{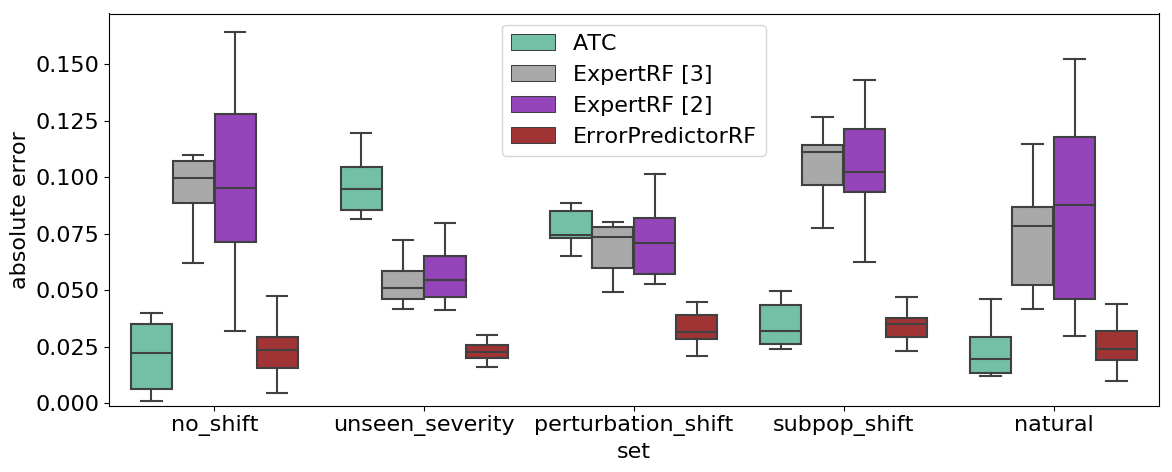}  
}
\subfloat[jsbach\_chorals]{
  \includegraphics[width=.45\linewidth]{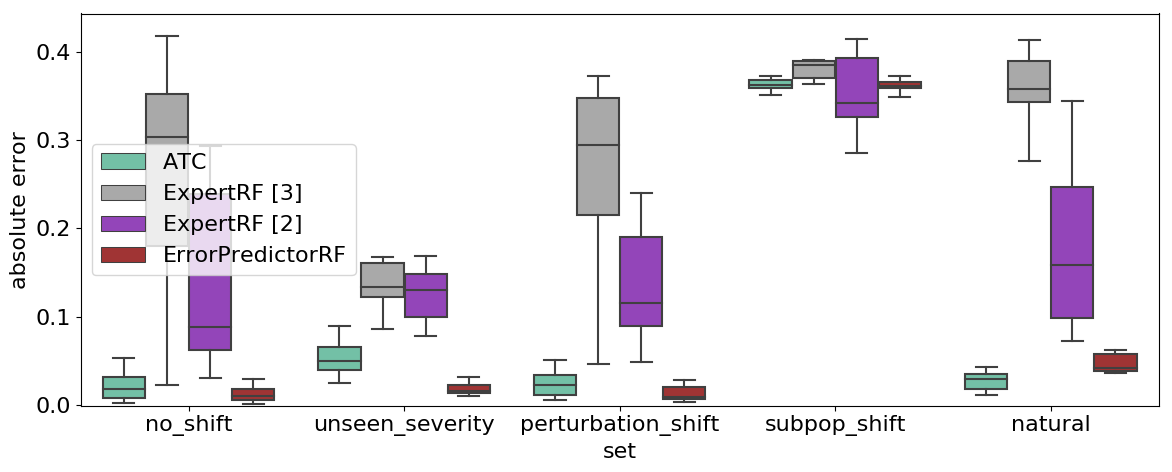}  
}\\
\subfloat[SDSS]{
  \includegraphics[width=.45\linewidth]{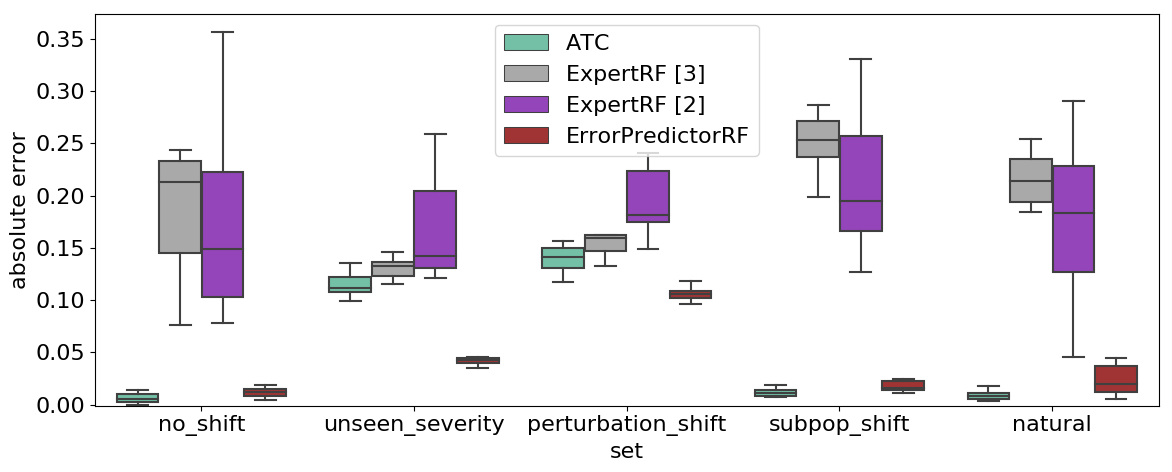}  
}
\subfloat[video\_games]{
  \includegraphics[width=.45\linewidth]{figures/video_games/abs_error_paper}  
}
\caption{Absolute prediction error of performance predictors on all datasets.}
\label{fig:all_datasets_abs_err}
\vskip -0.2in
\end{figure*}

The absolute prediction error of performance predictors on all datasets is reported in Figures \ref{fig:all_datasets_abs_err}.

\end{document}